# An Attention-Driven Heterogeneous GNN Model for Credit Card Fraud Detection

**KATHIRESAN JAYABALAN [1], SETHURAMAN RADHAKRISHNAN [2]**

[1] Research Scholar, Department of Computer Science and Engineering, Sathyabama Institute of Science and Technology, Chennai, 600119, Tamil Nadu, India
[2] Associate Professor, Department of Computer Science and Engineering, Sathyabama Institute of Science and Technology, Chennai, 600119, Tamil Nadu, India

E-mail: [1] kathiresan.jayabalan@gmail.com, [2] sethuraman.cse@sathyabama.ac.in
***Abstract -*** *The global transition to a cashless economy has placed credit cards as the key element of digital transactions, acclaimed for their easy use, speed, and acceptance in most places. However, the growing dependence on this payment method has led to an escalation of the risks associated with credit card (CC) fraud. Detecting this type of fraud is a difficult task because the patterns are constantly changing, there is a data imbalance, and it is necessary to identify the legitimate transactions and the fraud ones at the same time. This study addresses this challenge by proposing a credit card fraud detection (CCFD) framework using a data balancing technique and a deep learning (DL) model. The proposed fraud detection model is trained and evaluated by collecting the dataset called Credit Card Fraud Detection from the Kaggle repository. As the dataset is highly imbalanced, we utilized the Synthetic Minority Oversampling Technique (SMOTE)-Tomek technique to balance the dataset. Further, the balanced dataset is classified using the Heterogeneous Graph Neural Network (HGNN) model. The HGNN model represent various transactions using a heterogeneous graph architecture and by using an attention-based message passing technique, it managed to consider the complex relationships, time factors, and user behavior. The integration of SMOTE-Tomek in the model further boosted its capacity to identify fraudulent transactions, while lowering the rate of false positives. The HGNN model attained a 99.97% accuracy, a 99.48% F1-score, a 99.15% precision, and a 98.97% recall. The findings indicates that this model is effective and can be applied to real-world CCFD scenarios.*

## 1. Introduction

Rapid technological advancements and the convenience of electronic services have resulted in a growing number of credit card transactions. Hence, security problems, including credit card fraud, which has become a serious issue for both the banks and the clients, have been on the rise along with the transactions [1]. Fraud is a criminal act and at the same time a civil violation. In recent years, financial frauds in banks, credit cards, financial statements, insurance, corporate financial activities, money laundering, and stock market have brought the greatest concern and attention.

Information technology (IT) advancements have a significant effect on the banking sector. Nowadays, all banking system transactions are represented mostly by credit cards and online net banking, but they are also more vulnerable to the new threats. The use of credit cards and other online payment methods has increased tremendously in recent years. However, this is also one of the reasons for the increase in credit card frauds. Different types of credit card frauds occur. The actual theft of a credit card is the most common type. The theft of secret information of the credit card is the second type of cyber fraud. Another fraud is committed when the CC information is used without the consent of the card holder in the online transactions [2].

Credit card fraud is showing an increase with the development of technologies. It is estimated that the loss due to credit card fraud was amount to 37 billion USD in 2024, showing a rise of 19% over the previous year. Most of this is due to card-not-present (CNP) transactions, that is when the card is not physically swiped such as online transactions. Unfortunately, almost 80% of e-commerce credit card fraud losses, reflecting the difficulties in verifying transactions in remote payment scenarios, are caused by CNP fraud [3].

A taxonomy of financial frauds at a high level is shown in Figure 1, which groups the frauds into three principal categories: bank fraud, corporate fraud, and insurance fraud. Bank fraud includes various illegal acts like mortgage fraud, credit card fraud, and money laundering fraud, which often take advantage of the vulnerabilities in banking and payment systems. Corporate fraud is a term that encompasses various deceptive operations, such as fraud in financial statements, and frauds in securities and commodities, whereby either persons or companies are engaged in tampering with financial data or manipulating the market for illegal profits. Insurance fraud is shown through the examples of healthcare insurance fraud and automobile insurance fraud which expose the schemes of false claims and misrepresentation in the insurance transactions. In general, the figure gives a clear view of the various kinds of frauds, showing that the area of financial fraud is large and that each type of fraud requires a different method of detection [4].

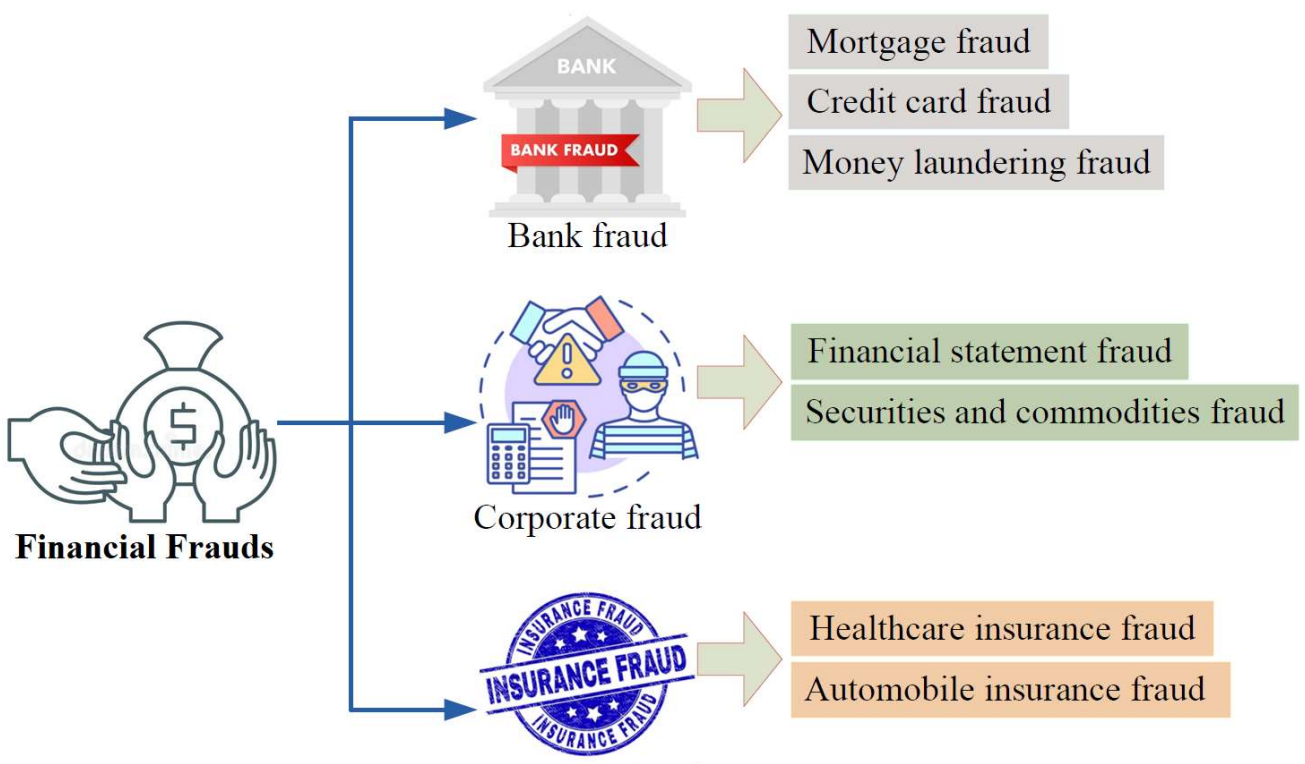

**Fig. 1 Financial Frauds Taxonomy**

Typically, fraud detection systems are based on analyzing customer data, which consists of online browsing, previous transactions, and patterns. These approaches utilize both data mining and data engineering techniques to processes raw events from active systems, such as logging data [5]. The artificial intelligence (AI) algorithms use data extraction and pattern analysis in classification methods to predict fraud behavior. Classification algorithms are an AI category that targets recognizing the class of a certain observation through a training dataset. The training of these algorithms involves the input of prior fraudulent observations for them to recognize the patterns of fraud. After that, the algorithms can classify a new customer on the website as soon as they get new input, for instance, if the new customer is similar to any of the previous ones [6]. Despite the ability of AI techniques to perform very well, there are still a few advantages and disadvantages in implementing them for CCFD [7].

With the digital economy expanding, there are growing complexities with fraud. There is definitively a requirement for more developed, robust, and adaptable detection models. The conventional manual detecting methods are not adequate anymore. Therefore, the financial sector is relying more on technologies like machine learning (ML) and DL to strengthen fraud detection and adaptability. These technologies can process high-dimensional data as well as identify complex fraud patterns and provide interpretable critical insights which are of great importance for compliance and trust in CCFD [8].

### *1.1. Problem Statement*

The financial sector has been experiencing an increase in fraudulent activities due to technological advancements and the overall economic growth that is visible in today's world, and the total cost of these activities to both financial institutions and consumers is in the range of hundreds of billions of dollars every year. The financial sector is one of the main targets of fraudsters who are constantly changing their methods to take advantage of the weaknesses of the existing prevention measures [9]. Among these offenses, we can find credit card fraud, fraudulent activities in the automobile and healthcare insurance sectors, insider trading, money laundering, fraud regarding commodities and securities. Independently, fraud prevention models are not sufficient to completely eradicate these types of crimes. Therefore, the existence of fraud detection frameworks to identify fraud activities that already have been carried out and the possible financial gain by doing so is more apparent than ever [10]. Thus, this study proposes an efficient CCFD framework using a DL model.

### *1.2. Research Novelty & Objectives*

This study's novelty is the development of an effective CCFD system that makes use of a HGNN to represent intricate transactional relations even in high class imbalance case. The proposed model, in contrast to traditional ML and DL methods that consider transactions as independent cases, can detect multi-entity and relational dependencies by means of heterogeneous graph creation together with attention-driven message passing. The application of SMOTE-Tomek balancing in a graph-learning pipeline not only improves the fraud detection sensitivity but also maintains the quality in terms of precision. Thus, this unified approach demonstrates to be a robust and generalizable solution for detecting real-time financial fraud.

The key objectives of this study are discussed as follows.

- To develop a fraud detection framework that integrates SMOTE-Tomek and HGNN for the precise CCFD.
- To model the complex transactional relations and multi-entity interactions through heterogeneous graph representation.
- To apply the CCFD dataset to train and evaluate the model.
- To address the high data imbalance in transactions of credit cards utilizing the SMOTE-Tomek sampling.
- To enhance the fraud detection accuracy by incorporating the attention-based message passing in the HGNN model.
- To assess the results based on parameters like precision, accuracy, specificity, F1-score, and recall.
- To compare the proposed framework to state-of-the-

art models, thereby proving its superiority over current fraud detection models.
- To conclude and discuss the advantages and limitations of the developed model with further improvements.

The paper is organized into the subsequent sections. The following section discusses the recent current works proposed on CCFD. The next section presents the modelling of the developed framework using SMOTE-Tomek and HGNN model. The experimentation analysis section is presented in Section 4, and finally, the research work is concluded with further works.

## 2. Literature Review

The most recent models that were designed for detecting CC fraud using various techniques have been reviewed in this section. Critically, the analysis of the analyzed models along with their advantages and disadvantages is presented in Table 1.

An ensemble ML model was proposed in [11] for finding the credit card fraud transactions. The SMOTE and Edited Nearest Neighbor (ENN) methods were incorporated to eliminate the problem of imbalanced data. The ensemble model consisted of Adaboost, K-nearest neighbor (KNN), random forest (RF), and a voting classifier, utilizing their strengths for better fraud detections. The superior performances of the ensemble method was demonstrated and the capability of the integration of multiple classifiers for increasing the accuracy of fraud detections was revealed.

In [12], a detection method was presented that employed anomaly detection algorithms for identification of CC fraud transactions. The system also used Principal Component Analysis (PCA) to reduce dimensionality but ensured that the essential information for accurate classification was still intact. Among the different models of fraud detections in credit card transaction, ABOD (Angle-Based Outliers Detections), CBLOF (Clusters-Based Local Outliers Factors), CD (Clusters-based Distances), COPOD (Clusters-based Outliers Detections), ECOD (Elliptic Envelopes-based Outliers Detections), LOF (Local Outliers Factors), IF (Isolation Forests), and HBOS (Histograms-based Outliers Score) were the ones selected for their effectiveness. Out of all the classifiers, the CBLOF classification showed the best performance.

The research in [13] made use of two unsupervised models, the Autoencoder (AE) and Generative Adversarial Networks (GAN), in an integrated manner as the Adversarial AE (AAE) for the prediction of credit card frauds. The GANs demonstrated the primary distributions of legitimate transaction and spotting patterns as deviations from the distributions learned, thus even eliminating the need for any fraud detection evidence. They were the AAEs, which further facilitated this process by integrating reconstructions-based training with adversarial regularizations and thus obtaining the latent spaces that were better organized and because of that improved the detection of anomalies.

The enhanced hybrid DL model proposed in [14] combined the capabilities of the Convolutional Neural Network (CNN) and the Recurrent Neural Network (RNN) for accurate detection of credit card frauds. The model successfully recognized the transaction patterns that were both sequential and reflectively spatial in making the anomaly detection process efficient with low rates of false negatives and false positives. The outcome of the experiments revealed the CNN-RNN model having the highest performance. These results highlighted the real-time detection of the fraud as a strongpoint of the CNN-RNN model, which also delivered excellent performance in classification with a minimum of false positives and an extensive coverage of anomalies.

The research in [15] analyzed the methodological integrity of the algorithms used to detect credit card fraud and showed that the complexities involved in the algorithms could easily be obscured by typical shortcomings in the assessment. The research has indicated that techniques were revealed by four major problems: improper sequencing of preprocessing steps which leads to unintentional data leaks; lack of agreement among researchers regarding the methodological descriptions; limited temporal validations of transaction information; and metrics distortion with the optimizations of recall at the cost of precision. The assessment which was done through a basic multilayer perceptron (MLP) baseline, in contrast to many recent works that have reported almost perfect recall (over 99%) through complex deep models, still offered modest but reliable measures and points out the widespread overestimation that was characteristic of defective pipelines.

The ML-based fraud detection system developed in [16] was able to discover the class imbalance using preprocessing methods such as SMOTE-ENN and further used an AE for dimensionality reduction and TOPSIS for feature selection. A predictive accuracy boost was achieved by stacking support vector machine (SVM), K-nearest neighbor (KNN), and extreme learning machine (ELM) within an ensemble algorithm. The particle swarm optimizer (PSO) was effective in tuning the parameters of ELM leading to better model convergence and generalization. The system's performance was demonstrated by the precision with which it could detect frauds with low rate of false positives.

The research in [17] considered the use of real credit card data sets for the assessment of logistic regressions, decision trees, and random forests (RF) approaches paying attention to the class imbalance and improving the accuracy of prediction. The focus loss was implemented in a DL approach to boost the detection power. The class imbalance issue was handled using SMOTE technique and hyperparameters tuning were done to set the approach accordingly. The experiments showed that the RF model achieved the overall better performance while the Light Gradient Boost model showed the highest precision underlining its ability to lower false positives.

In [18], a hybrid methodology was proposed for detecting CC fraud which combined Quantum Machine Learning (QML) with classical and meta-heuristic methods for feature selection. The feature selection methods were comprised of the K-Best traditional methods as well as ASO, ACO, and PSO meta-heuristic techniques, and on the basis of the Variational Quantum Classifier (VQC) each of them was thoroughly evaluated for classification. The accuracy of the PSO+VQC pair was 94.54%, thus proving the efficacy of using meta-heuristic methods together with VQC in handling complex, high-dimensional data for fraud detection.

The research in [19] implemented and evaluated several generative models including standard AE, GAN, Variational AE (VAE), and an AE-GAN model for the purpose of detecting credit card frauds. These models created synthetic fraud instances for balancing the data set and, consequently, improved the learning of the method. A real-time credit card dataset was analyzed, and the conclusion was that the use of generative algorithms had a significantly positive impact in reducing the imbalance problem usually present in the detection of fraudulent transactions.

In [20], a model for the detections of CC frauds was developed which combined the hybrid Big Bang-Big Crunch (BB-BC) with Cuckoo Search (CS) method for the feature selection. The method utilized the BB-BC method for the local exploitations of one solution and CS for the global explorations of multiple solutions. The CS method relied on Levy flights to support the BB-BC agents escaped from being stuck and converging early. After selection of features, Deep CNN (DCNN) and EDCNN were used to increase accuracy through classification. The method performed the best with DCNN at 94.59% and with EDCNN at 95.61%.

A hybrid CCFD framework that integrated CNNs, Long Short-Term Memory networks (LSTM), and an attention mechanism was presented in [21]. The hybrid model's parts were deliberately built to get different features of the behavior: CNNs focus on the spatial features, LSTMs for the temporal sequences, and the attention mechanism was to accentuate most important features. The combined CNN-LSTM-Attention model improved detection of fraud by simultaneous treatment of temporal and spatial characteristics of data coupled with the dynamically highlighting of essential components. The results proved the model's excellent capability of identifying fraudulent activities with high accuracy and reliability.

In [22], a hybrid feature selection methodology was developed that would enhance the performance of ML models in detecting credit card fraud. The hybrid feature selection methodology combined Random Forest Importance (RFI), Information Gain (IG), and Pearson Correlation to reach the best selection of features. The setup involved state-of-the-art ML models, like AdaBoost, CatBoost, Extra Trees (ET), XGBoost (XGBC), RF, as well as an ensemble voting mechanism, thus resulting in higher accuracy in detection.

In [23], an ensemble DL methodology was proposed that combined CNN, Gated Recurrent Units (GRU), and MLP with the SMOTE-ENN to handle the class imbalances and improve the precision of detections. The approach utilized CNN for feature extraction, GRU for analyzing sequential transactions, and MLP as a meta-learning algorithm in a stacking architecture. The ensemble method achieved a remarkably high detection rate and enhanced generalization.

An ensemble ML approach was proposed in [24] for CCFD with enhanced safety as a result. The ensemble model integrated methodologies of AdaBoost, logistic regression, and RF to efficiently distinguish between legitimate and fraud transactions. The ensemble method has the advantage of each model to develop an adaptable and robust system for the real-time detection of the evolving fraud activities. The findings demonstrated the approach's capability to keep very low levels of both false negatives and false positives, thus, making it a reliable choice for fraud detection.

The research in [25] proposed a Temporal Heterogeneous Graphs Contrastive Learning (TH-GCL) methodology for the purpose of CCFD combining heterogeneous graphs representation, temporal modelling, and contrastive learning. The approach was not only about the creation of temporal heterogeneous graphs that included the complex multi-entity relations of financial transaction networks but also about the use of time-aware GNN models with attention mechanism for achieving thoroughly representation and finally, through dual-view contrastive learning, resilience enhancement in sparse labelling situations. Thus, the model differentiated the legitimate from the fraud transactions and at the same time delivered stable results.

**Table 1. Critical Analysis of Reviewed Research Models**

| Ref. | Models | Applications | Advantage | Disadvantage |
|---|---|---|---|---|
| [11] | Ensemble ML (AdaBoost, KNN, RF, | Transaction-level fraud detection | Improved detection accuracy through ensemble learning; | High computational complexity; limited ability to |
|  | Voting) with SMOTE-ENN |  | effective imbalance handling. | model temporal dependencies. |
| [12] | PCA + Anomaly Detection (CBLOF, LOF, IF, HBOS, etc.) | Unsupervised fraud detection | Does not require labelled fraud data; CBLOF shows strong anomaly detection. | Limited adaptability to evolving fraud patterns; sensitive to parameter settings. |
| [13] | Adversarial Autoencoder (GAN + AE) with explainability | Unsupervised fraud prediction | Learns normal transaction distribution; enhanced interpretability using Shapley values. | Training instability; computationally expensive. |
| [14] | Hybrid CNN–RNN | Real-time fraud detection | Captures spatial and sequential patterns; low false positives. | Requires extensive training data; limited explainability. |
| [15] | Methodological evaluation using MLP baseline | Evaluation integrity analysis | Highlights data leakage and metric inflation issues; realistic performance assessment. | Does not propose a novel detection model. |
| [16] | Stacked Ensemble (SVM, KNN, ELM) with SMOTE-ENN, AE, PSO | Transaction fraud detection | Optimized feature selection and model parameters; reduced false positives. | Increased pipeline complexity; higher training cost. |
| [17] | ML (LR, DT, RF) + DL with Focal Loss | Imbalanced fraud detection | Improved recall and precision trade-off; RF shows strong performance. | Limited modelling of relational transaction dependencies. |
| [18] | Quantum ML (VQC) with Meta-heuristic Feature Selection | High-dimensional fraud detection | Demonstrates potential of QML; effective feature optimization. | Scalability and hardware constraints; early-stage applicability. |
| [19] | Generative Models (AE, GAN, VAE, AE-GAN) | Data imbalance mitigation | Effective synthetic fraud sample generation; improves minority learning. | Risk of generating unrealistic samples; training instability. |
| [20] | BB-BC + Cuckoo Search with DCNN/EDCNN | Feature-optimized fraud detection | Enhanced accuracy via hybrid optimization; strong feature selection. | Computationally intensive; limited interpretability. |
| [21] | CNN-LSTM with Attention | Temporal and spatial fraud detection | Simultaneous modelling of sequential and spatial features; high accuracy. | Model complexity; high resource requirements. |
| [22] | Hybrid Feature Selection + Ensemble ML | Feature-driven fraud detection | Improved model efficiency and accuracy; reduced feature redundancy. | Performance depends on feature quality; lacks temporal modelling. |
| [23] | Ensemble DL (CNN-GRU-MLP) with SMOTE-ENN | Sequential fraud detection | Strong generalization; effective imbalance mitigation. | Complex stacking architecture; difficult deployment. |
| [24] | Ensemble ML (AdaBoost, LR, RF) | Real-time fraud detection | Robust and flexible; low false positives and negatives. | Limited deep feature learning capability. |
| [25] | TH-GCL | Network-based fraud detection | Captures multi-entity relations and temporal dynamics; robust under sparse labels. | High computational cost; complex graph construction. |

## 3. Materials and Methods

This study developed a novel fraud detection framework that is developed based on a systematic workflow for the appropriate detection of credit card fraud utilizing a HGNN, which is shown in figure 2. The first step in the model is the collection of CCFD dataset which is open-access, and then it is performed through a process of data cleaning and preprocessing. This process includes data integrity validation, noise handling, feature scaling (mostly for transaction amount and time), and preparing graph-based learning suitable features. The preprocessed dataset is then split into training and testing sets through an appropriate data-splitting method to prevent any information leakage. Due to the imbalanced nature of the dataset, data balancing is applied solely on the training data through the SMOTE-Tomek technique, which not only oversamples the minority fraud instances but also removes the ambiguous majority samples thus improving the class separability.

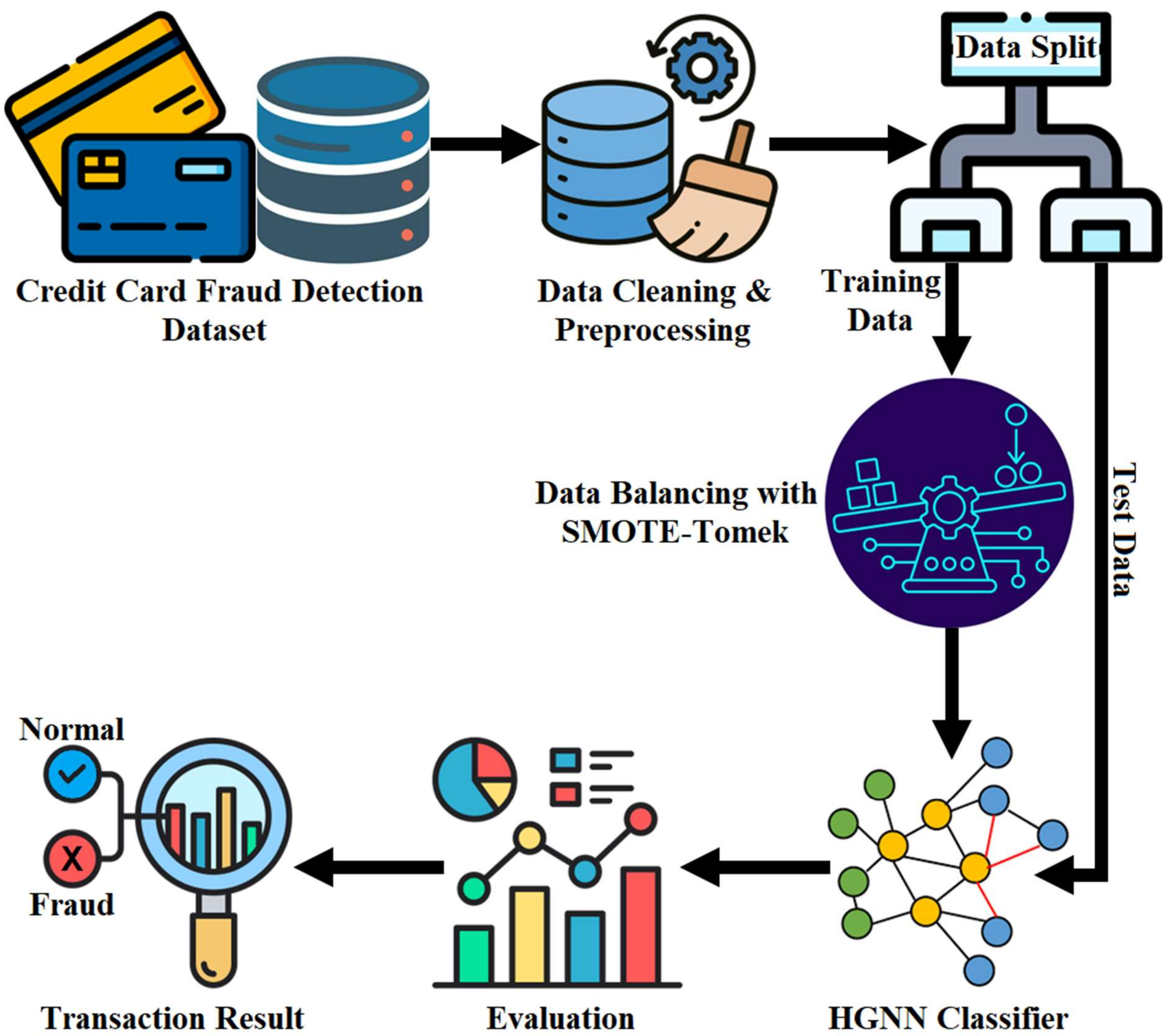

**Fig. 2 Workflow of the Developed Research Model**

The balanced training data is then transformed into a heterogeneous graph structure where transactions along with their respective contextual entities are represented as nodes that are interconnected through significant relations. The graph is fed to the HGNN classifier which carries out type-aware message passing and aggregation to detect complex relational patterns showing fraudulent behavior. The model which has been trained is assessed against the unknown test data, and the performance of the model is rated with the help of standard metrics for evaluation, which also indicates the effectiveness of the model under class imbalance conditions. In the end, the transactions are categorized as normal or fraudulent depending on the classification output, showcasing the proposed framework's practical applicability to real-world CCFD scenarios.

### *3.1. Dataset Details*

The Credit Card Fraud Detection dataset utilized in this study is obtained from the ULB Machine Learning Group, with its description accessible at the Kaggle repository (https://www.kaggle.com/datasets/mlg-ulb/creditcardfraud). This dataset comprises credit card transaction details carried out by card holders throughout Europe in 2013 September. From the 2,84,807 evaluated transactions, 492 are identified as fraudulent, as indicated in Table 2.

**Table 2. Clinical Information of the Dataset**

| Total Transactions | Features | Classes | Real Transactions | Fraudulent Transactions | Attributes | Description |
|---|---|---|---|---|---|---|
| 2,84,807 | 31 | 2 | 2,84,315 | 492 | Class | Target variable where 1 denotes fraud and 0 denotes legitimate transaction. |
| | | | | | Amount | Total transaction amount; useful for cost-sensitive learning. |
| | | | | | Time | Seconds elapsed between the initial transaction and all the subsequent transactions. |
| | | | | | V1, V2, …, V28 | Principal components obtained using PCA. |

The dataset is highly imbalanced due to the infrequent incidence of scams, which constitute just 0.172% of the total transactions. Privacy considerations necessitated the alteration of PCA, resulting in the incorporation of numerous anonymized features within the dataset. These features predominantly comprise numerical values. The sole features excluded from anonymization are "Amount" and "Time". The column with the term "time" in the data set denotes the duration in seconds from the initial transactions for all the subsequent transactions. "Amount" denotes the total amount of the transactions. The rest of the features were designated as V1, V2, …, V28. The "Class" term signifies the transaction category, with '1' denoting a fraud transaction and '0' indicating non-fraud transaction. The term "Amount" denotes the financial value linked to the transactions [26]. This data set was frequently utilized for assessing detection of anomaly systems, especially in frauds detection contexts. The imbalanced characteristics and anonymized features make conventional models ineffective in identifying fraud accurately.

### 3.2. Preprocessing

The performances of the developed HGNN-based CCFD framework relies heavily on a strong preprocessing pipeline, particularly because of the dataset's highly unbalanced nature and characteristics of transactions. The purpose of the preprocessing stage is to clean the dataset, standardize the distributions of features, handling class imbalance, and convert the tabular data into a graph format.

#### *3.2.1. Data Cleaning*

At first, the dataset is analyzed for missing data, repeated records, and variations in the presentation of features. Even though the dataset is accessible to the public and already cleaned, additionally analyzed to make sure of the data quality before training the model. All the duplicate transactions are eliminated to prevent the learning process from being affected by bias.

Let $D = \{x_i, y_i\}_{i=1}^{N}$ denotes the transaction dataset. Here, $x_i \in \mathbb{R}^d$ indicates the feature vector and $y_i \in \{0,1\}$ denotes the class label. The dataset after cleaning is attained as follows.

$$D_{clean} = D / \{x_i | x_i = x_j, i \neq j\} \quad (1)$$

This process makes every transaction contributes individually for the learning process [27].

#### *3.2.2. Feature Scaling & Normalization*

The dataset includes 28 PCA-transformed and already normalized features $V_1 - V_{28}$, with two raw features like Time and Amount. To make sure that each feature has the same effect in the process of graph learning, additional scaling is performed.

The transaction Amount feature usually includes extreme values, which can affect the learning of the model. Therefore, robust scaling is utilized as follows:

$$Amount' = \frac{Amount - median(Amount)}{IQR(Amount)} \quad (2)$$

Here, $IQR$ is the indication of interquartile range. This scaling method minimizes the impact of outliers but keeps the information of the relative size intact.

The Time attribute was normalized using the Min-Max normalization using the following equation [28].

$$Time' = \frac{Time - Time_{min}}{Time_{max} - Time_{min}} \quad (3)$$

This normalization process stabilizes the temporal variations and enhances the convergence in HGNN model.

*3.2.3. Temporal Feature Engineering*

Fraud activities show a very strong correlation with time. To represent such dynamics, the transaction timestamps are used to create the temporal features. The transactions are divided into predefined time intervals, and the time gaps between the transactions are calculated as follows:

$$\Delta t_i = Time_i - Time_{i-1} \tag{4}$$

Temporalizing such representations can allow the model to learn the normal and abnormal behavior associated with fraudulent transactions.

*3.2.4. Data Split*

For preventing the leakage of information and simulating the deployment case, the dataset is split into training and testing sets by employing a time-aware splitting technique:

$$D_{train} = \{x_i | Time_i \leq T_{split}\}, D_{test} = \{x_i | Time_i > T_{split}\} \tag{5}$$

This makes sure that test data transactions will not be involved in the model training process. In this study, 70% of the data is allocated for training and the other 30% is set for testing.

*3.2.5. SMOTE-Tomek*

The dataset displays a significant imbalance in its classes as only 0.172% of the total samples are fraudulent transactions. To address this problem, the SMOTE–Tomek hybrid resampling technique is used only on the training data.

By implementing the SMOTE methodology, it generates the synthetic fraud samples by interpolating between the minority class instances using the following equation.

$$x_{new} = x_i + \lambda(x_{nn} - x_i), \;\; \lambda \in (0,1) \tag{6}$$

Here, the variable $x_{nn}$ was one of the k-NN of $x_i$.

The Tomek links detects the overlapping majority-minority sample pairs using the following equation.

$$(x_i, x_j) \; is \; Tomek \; link \;\;\; if \; d(x_i, x_j) < d(x_i, x_k) \tag{7}$$

Here, the variables $x_i$ and $x_j$ belong to different classes. The majority class instance is eliminated to improve class separability [29].

*3.2.6. Graph-Oriented Transformation*

Heterogeneous graph learning is made possible by the conversion of the preprocessed tabular data into a heterogeneous graph $G = (V, E)$. It identifies various types of nodes, such as transaction nodes, time-window nodes, and amount-bin nodes. Edges are determined by temporal proximity, amount similarity, and transactional relationships. The node feature vectors are represented as follows:

$$h_v^{(0)} = \phi(x_v) \tag{8}$$

Here, $\phi(\cdot)$ identifies the feature projection function. Such a transformation will enable the heterogeneous graph to track those relational dependencies.

Categorical and structural features are one-hot encoded, and continuous features are fed through an MLP (multilayers perceptron) to map into a common embedding space:

$$h_v = \sigma(W x_v + b) \tag{9}$$

Here, $W$ and $b$ were the learnable parameters and $\sigma$ was the nonlinear activation function.

This preprocessing stage guarantees data integrity, mitigates class imbalance, captures temporal transaction behavior, and allows for graph-based relational modeling. Together, these steps improve the discriminative ability and stability of the HGNN classifier in detecting fraudulent credit card transactions.

### ***3.3. HGNN Modelling***

This study proposed a HGNN-based classification framework to effectively detect credit card fraud by representing its complicated relational structure and transaction dependencies. The traditional ML and the DL models of the same type routinely process transactions one by one, while the HGNN, on the other hand, explicitly depicts the interactions between various heterogeneous entities and thus opens a path leading to the robust learning of the fraud-related behaviors [30].

The overall architecture of the proposed CCFD model based on HGNN is depicted in Figure 3. The first step consists of supplying the model with preprocessed transaction data containing numerical attributes that have been scaled, class distributions that have been balanced, and temporal features. The data is subsequently represented as a heterogeneous graph which is made up of various node types like transaction nodes, time-window nodes, and amount-bin nodes connected via edges that are specific to the relations and represent the proximity in time and the similarity in amount.

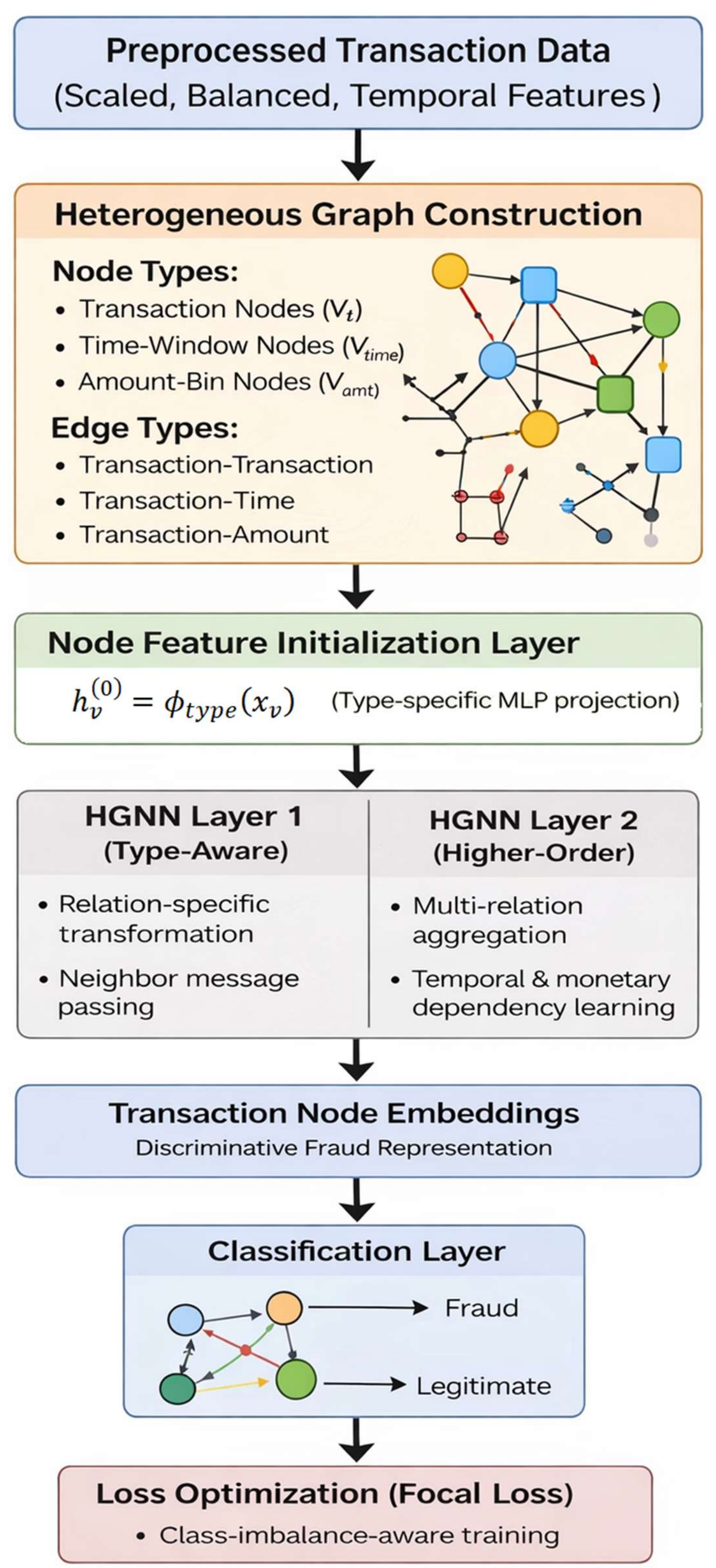

**Fig. 3 Architecture of Proposed HGNN Model**

Different node categories are mapped to a common latent feature space by means of a particular node feature initialization layer for each type. After that, the HGNN applies the graph to the multiple type-aware and higher-order graph convolution layers, thus allowing for message passing that is specific to a relation and the aggregation of multiple relations to learn intricate dependencies of transactions.

The transaction node embeddings produced from this process represent distinct fraud-related patterns and are forwarded to a classification layer which, in turn, results the probability of each transaction being either fraudulent or legitimate. In the end, a focal loss-based optimization module is utilized in training to handle the critical class imbalance and to increase the model's sensitivity to infrequent fraudulent transactions, thus leading to an overall enhancement in detection performance.

The conversion of transactional data into a heterogeneous graph is done in the following manner:

$$G = (V, E, T_v, T_e) \tag{10}$$

Here, $V$ indicates the set of nodes, $E$ indicates the set of edges, $T_v$ denotes the node types, and $T_e$ denotes the edge types. The graph consists of several different types of nodes that illustrate the various transaction entities: Transaction nodes are represented by $v_t$, the time-window nodes are represented by $v_{time}$, and amount-bin nodes are represented by $v_{amount}$. Each node $v \in V$ is associated with a type $\tau(v) \in T_v$.

Meaningful relationships between nodes of different types are represented by edges:

- Transaction–Time edges show the connection in time.
- Transaction–Amount edges display the similarity in money.
- Transaction–Transaction edges depict the closeness in time.

Formally, the edge is expressed as follows in the given equation.

$$e = (v_i, v_j, r) \tag{11}$$

Here, $r \in T_e$ indicates the relation type. These types edges allow the HGNN model to perform the relation-aware message passing.

Each node $v$ was initialized with a feature vector using the following equation.

$$h_v^{(0)} = \phi_{\tau(v)}(x_v) \tag{12}$$

Here, $x_v$ indicates the raw feature vector and $\phi_{\tau(v)}(\cdot)$ is a type-specific feature projection function, which is implemented by applying an MLP. This allows all heterogeneous nodes are embedded into a unified latent space.

HGNN carries out message passing through the aggregation of data from adjacent nodes, taking into account the types of both nodes and edges. As for a node $v$, the calculation of the message from a neighboring node $u$ through relation $r$ at layer $l$ is done as follows:

$$m_{u \to v}^{(l)} = W_r^{(l)} h_u^{(l)} \tag{13}$$

Here, $W_r^{(l)}$ is the relation-specific transformation matrix.

The aggregated message for a node $v$ is then obtained by the following equation.

$$m_v^{(l)} = \sum_{r \in \tau_e} \sum_{u \in N_r(v)} \alpha_{uv}^{(r)} m_{u \to v}^{(l)} \tag{14}$$

Here, $N_r(v)$ indicates the neighbors of $v$ under relation $r$, and $\alpha_{uv}^{(r)}$ indicates the attention weight assigned to neighbor $u$.

To emphasize informative neighbors and suppress noisy interactions, an attention mechanism is employed using the following equation.

$$\alpha_{uv}^{(r)} = \frac{\exp\left(\sigma\left(a_r^\top \left[h_u^{(l)} || h_v^{(l)}\right]\right)\right)}{\sum_{k \in N_r(v)} \exp\left(\sigma\left(a_r^\top \left[h_k^{(l)} || h_v^{(l)}\right]\right)\right)} \tag{15}$$

Here, $a_r$ is the learnable relation-specific attention vector, $||$ indicates the vector concatenation, and $\sigma(\cdot)$ indicates a non-linear activation function. This makes the model to dynamically prioritize the fraud transactional interactions.

The node embeddings are updated utilizing the aggregated messages as represented in the following equation.

$$h_v^{(l+1)} = \sigma\left(W_0^{(l)} h_v^{(l)} + m_v^{(l)} + b^{(l)}\right) \tag{16}$$

Here, $W_0^{(l)}$ and $b^{(l)}$ were the learnable parameters. The multiple HGNN layers were stacked to capture high-order relational dependencies.

After $L$ HGNN layers, the final embeddings of transaction nodes were utilized for classification.

$$\hat{y}_v = softmax\left(W_c h_v^{(L)} + b_c\right) \tag{17}$$

Here, the variables $\hat{y}_v \in \mathbb{R}^2$ indicates the predicted class probabilities (legitimate or fraud), $W_c$ and $b_c$ were the classifier parameters.

Given the severe class imbalance, the focal loss was applied to focus learning on minority fraud samples using the following equation.

$$L_{FL} = -\sum_{i=1}^{N} \alpha(1-p_i)^{\gamma}\log(p_i) \quad (18)$$

Here, the $p_i$ indicates the predicted probability of the true class, $\alpha$ balances the class importance, and $\gamma$ controls the focusing strength [31].

The HGNN is trained in an end-to-end manner utilizing the Adam optimizer. All parameters are updated through backpropagation, which includes type-specific transformations, attention weights, and classifier layers. The training process uses balanced training data, while evaluation is conducted on the unseen test data to provide an unbiased assessment. The proposed HGNN framework successfully detects complicated fraud patterns that are not easily discoverable through flat or sequential models by combining heterogeneous entities, temporal context, and relation-aware attention. Learning the relationships between transactions greatly improves fraud detection, especially in case of evolving and sparse fraud behaviors.

The pseudocode for this developed research model is presented in the following.

**Algorithm: HGNN-Based Credit Card Fraud Detection**

*Input: Credit card transaction dataset D; Class labels (0: Legitimate, 1: Fraud) y; Train–test split ratio (70:30) r.*
*Output: Predicted class labels ŷ*
*Initialize*
*Load dataset D*
*Data Cleaning*
*Remove duplicate transactions from D*
*Verify missing values and data consistency*
*Apply Robust Scaling to Amount feature*
*Apply Min–Max normalization to Time feature*
*Compute inter-transaction time gap Δt*
*Segment transactions into fixed time windows*
*Split D into training set D_train and test set D_test based on Time*
*Apply SMOTE to generate synthetic fraud samples in D_train*
*Apply Tomek Links to remove overlapping majority samples*
*Define node types:*
  *Transaction nodes*
  *Time-window nodes*
  *Amount-bin nodes*
*Define edge types:*
  *Transaction–Transaction*
  *Transaction–Time*
  *Transaction–Amount*
*Construct heterogeneous graph G(V, E)*
*Node Feature Initialization*
*For each node v ∈ V do*
  *Initialize h_v⁰ = φ_type(x_v)*
*End for*
*HGNN Training*
*For each HGNN layer l = 1 to L do*
  *For each node v ∈ V do*
  *Aggregate relation-aware messages from neighbors*
  *Update node embedding h_vˡ using attention mechanism*
*End for*
*End for*
*Extract final embeddings of transaction nodes*
*Compute class probabilities using Softmax*
*Compute focal loss to handle class imbalance*
*Update model parameters using Adam optimizer*
*Evaluate model on D_test*
*Return predicted labels ŷ*
*End*

**Table 3. List of Proposed HGNN Model's Hyperparameters**

| Category | Hyperparameter | Value |
|---|---|---|
| Data Preprocessing | Train–Test Split Ratio | 70:30 (%) |
| | Scaling Method (Amount) | Robust Scaler |
| | Scaling Method (Time) | Min–Max Normalization |
| | Imbalance Handling | SMOTE–Tomek |
| | SMOTE Neighbors (k) | 5 |
| Graph Construction | Transaction Node Type | Transaction |
| | Time-Window Size | 1 hour |
| | Amount Bin Count | 10 |
| | Edge Types | 3 |
| HGNN Architecture | Number of HGNN Layers | 2 |
| | Hidden Dimension Size | 64 |
| | Attention Mechanism | Relation-aware attention |
| | Activation Function | ReLU |
| Training Parameters | Optimizer | Adam |

| | Learning Rate | 0.001 |
|---|---|---|
| | Batch Size | 256 |
| | Number of Epochs | 100 |
| | Weight Decay | 0.0001 |
| Loss Function | Loss Type | Focal Loss |
| | Focal Loss α | 0.25 |
| | Focal Loss γ | 2.0 |

The pseudocode highlights the complete workflow of the HGNN-based CCFD system, which starts from data preprocessing and imbalance handling through heterogeneous graph creation, model training, and evaluation. Time-aware data splitting and SMOTE-Tomek resampling techniques applied together guarantee that the learning is done in a realistic way even under a severe class imbalance situation, whereas different node and edge definitions allow the modelling of complicated transactional relationships effectively.

As shown in Table 3, the selected hyperparameters like a two-layer HGNN with a hidden dimension of 64, relation-aware attention, and ReLU activation make the model both expressive and computationally efficient. The usage of the Adam optimization, 0.001 learning rate, and focal loss during training leads to better convergence stability and makes the model more sensitive to rare fraud cases. Overall, the pseudocode along with the hyperparameter settings guarantees a robust, flexible, and imbalance-sensitive fraud detection model that is capable of being applied to the analysis of real-time financial transactions.

## 4. Experimentation Analysis

### *4.1. Experimental Setup*

The developed HGNN model was experimented with and evaluated utilizing the PYTHON 3.7.12 programming language. The experimentations are carried out on Google Colab Pro. The CCFD dataset was utilized for the experiment, which was split into 70% for training and 30% for testing. The experiments were performed on a system with an Intel Core i7 processor, 12 GB RAM, 1 TB SSD, and NVIDIA RTX 3080 GPU to ensure rapid data loading and training.

### *4.2. Evaluation Metrics*

The evaluation of HGNN model performance was conducted via standard parameters including accuracy, F1-score, precision, specificity, and recall, which together provided a comprehensive assessment of fraud detection reliability.

Accuracy: The efficiency of a model is assessed by its accuracy, determined by a ratio of TP and TN relative to all previous predictions. The following equation (19) is applied to compute the accuracy.

$$Accuracy = \frac{TP+TN}{TP+TN+FP+FN} \quad (19)$$

Precision: The TP rate, indicating the percentage of correctly predicted positive cases among all predicted instances, was utilized to evaluate the precision of the model. Equation (20) is utilized to assess the precision of the model.

$$Precision = \frac{TP}{TP+} \quad (20)$$

F1-Score: It is a precision and recall's harmonic mean and generates a balanced computation of the performance of a classifier. Equation (21) is applied to calculate the F1-score.

$$F1score = \frac{2 \times Precision \times Recall}{Precision+Recall} \quad (21)$$

Specificity: It is the measure of the model to accurately identify negative outcomes and is assessed by calculating the ratio of correctly classified TN. This statistic evaluates the classifier's potential to accurately identify and categorize fraud occurrences. The specificity of the model was computed by using equation (22).

$$Specificity = \frac{TN}{TN+} \quad (22)$$

Recall: It refers to the HGNN model's potential to precisely identify positive instances. It quantifies the proportions of true positive instances precisely recognized. It can also be referred to as recall. The following equation (23) is applied to compute the sensitivity of the model.

$$Recall = \frac{TP}{TP+FN} \quad (23)$$

True Positive (TP) indicates the total data records properly classified as fraud. True Negative (TN) is the total images correctly detected as not fraud. False Positive (FP) denotes the total count of non-fraud data incorrectly detected as fraud when they are not. False Negative (FN) denotes the overall count of fraud data inaccurately detected as non-fraud.

### *4.3. Performance Assessment*

The performance analysis of the developed research model is assessed based on both training and test datasets. The performances of the model were assessed utilizing common classification metrics like accuracy, specificity, precision, F1 score, and sensitivity. Both these performance assessments are tabulated individually in the following Tables 4 and 5.

**Table 4. Results of HGNN Model Using Training Dataset**

| Parameters | Legitimate | Fraud |
|---|---|---|
| Accuracy | 99.99 | 99.97 |
| Precision | 99.96 | 99.94 |
| F1-score | 99.97 | 99.95 |
| Specificity | 99.98 | 99.92 |
| Recall | 99.95 | 99.96 |

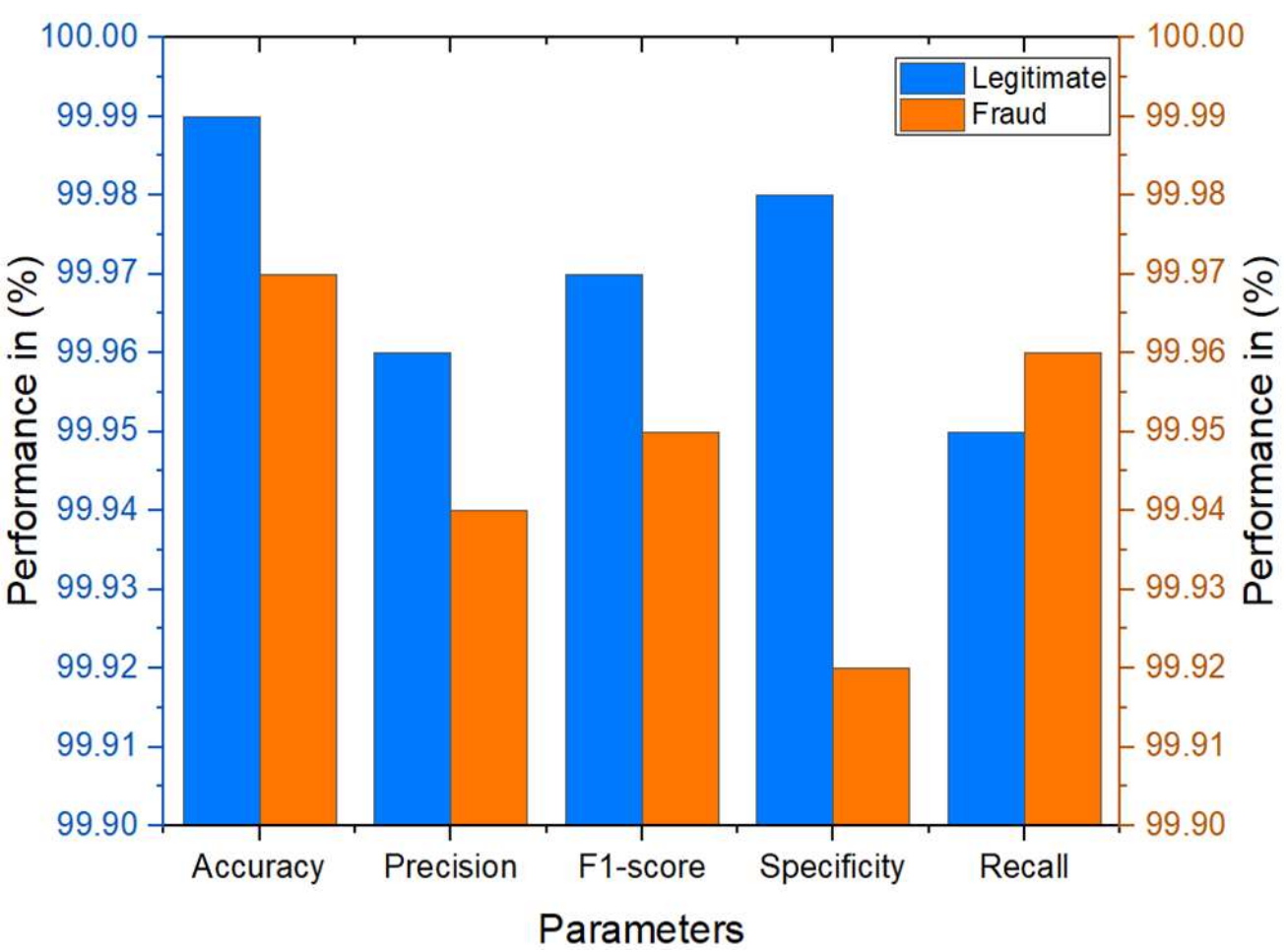

**Fig. 4 Graphical Illustration of Results on Training Data**

The proposed HGNN model's performance on the training dataset is shown in the Table 4 which indicates its effective classification ability of both legitimate and fraudulent transactions. Such high accuracy values like 99.99 and 99.97 for legit and fraud, clearly show that the model is superior at learning the main transaction patterns and with minimal misclassification while training. The precision numbers of 99.96% for legitimate transactions and 99.94% for fraud transactions indicate that the model has very few false positives across both classes, which is crucial, allowing the detection of fraud cases to prevent unnecessary alerts. The F1-scores of 99.97% and 99.95% have once again proved the excellent trade-off within recall and precision, which is the result of stable and consistent learning. Very high specificity values (99.98% for legitimate and 99.92% for fraud) demonstrate the model's strong capability in recognizing non-fraudulent transactions correctly and at the same time keeping a reliable fraud discrimination. The model's recall value was evident through the exceptional values of 99.95% for legitimate and 99.96% for fraud instances, which particularly marked the model as being effective in capturing fraudulent instances during training. Overall, it is confirmed that the HGNN model not only obtains strong convergence but also makes good use of different graph representations to capture the discriminative patterns in the training data. Figure 4 depicts the graphical illustration of the results on training data.

**Table 5. Results of HGNN Model Using Test Dataset**

| Parameters | Legitimate | Fraud |
|---|---|---|
| Accuracy | 99.98 | 99.97 |
| Precision | 99.82 | 99.15 |
| F1-score | 99.89 | 99.48 |
| Specificity | 99.87 | 99.64 |
| Recall | 99.91 | 98.97 |

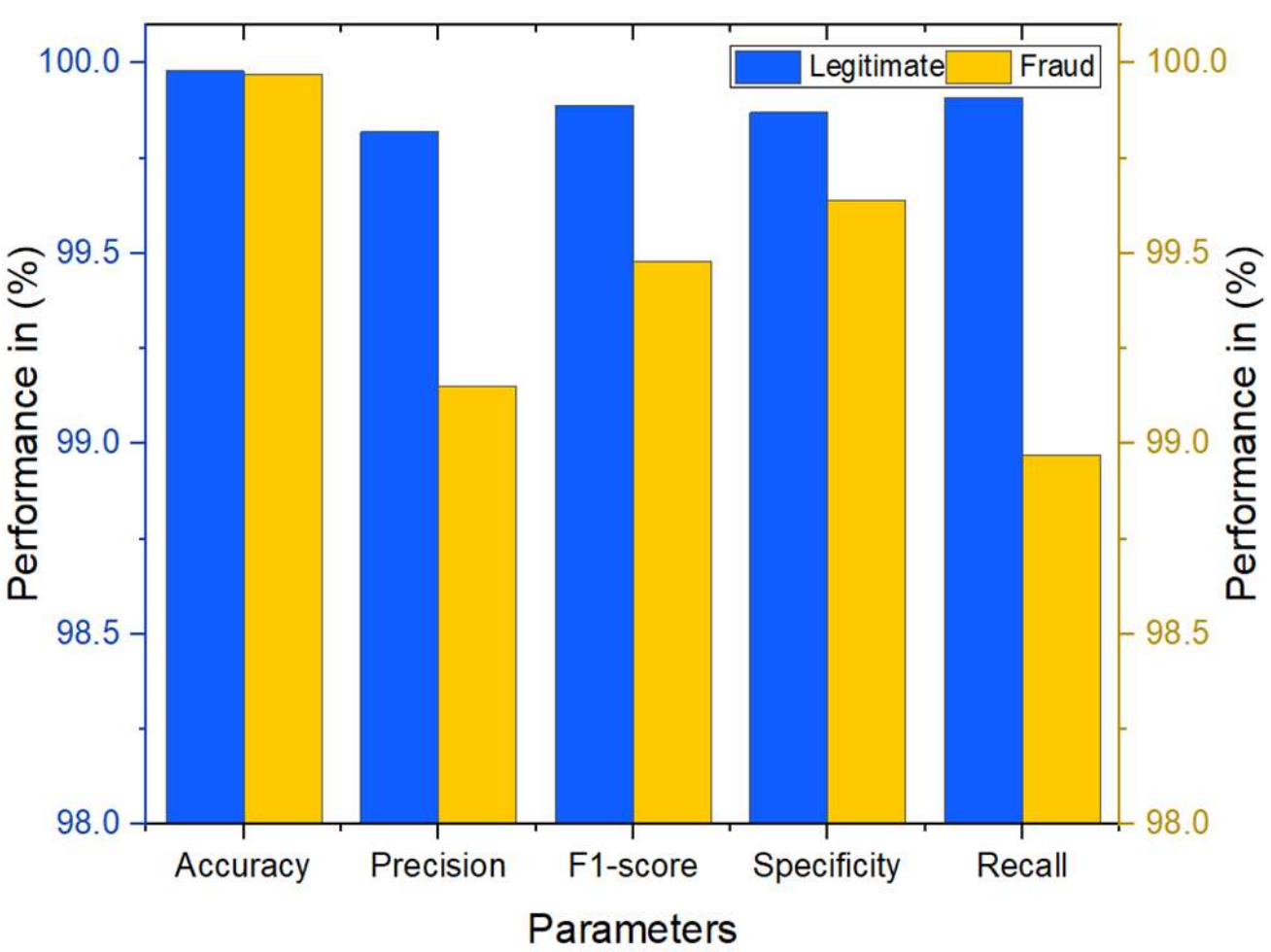

**Fig. 5 Graphical Illustration of Results on Test Data**

Table 5 tabulates the results of the developed HGNN model on the test dataset that was not seen, which indicates a superior ability of the model to generalize well in the classifying real-world data. The model consistently classified the data beyond the training set as evidenced by the exceptional accuracy rates of 99.98% for legitimate transactions and 99.97% for fraudulent ones. The precision scores of 99.82% for the legitimate and 99.15% for the fraud cases show that the HGNN has quite correctly reduced the false positives while at the same time identifying the fraud data accurately that were very important for minimizing the expenses in the financial systems. The F1 scores of 99.89% for non-fraud and 99.48% for fraud indicate a great balance of trade-off between precision and recall, thus confirming the reliability of the detection even in case of severe class imbalance. The high specificity scores of 99.87% for legitimate and 99.64% for fraud indicate the model's capability to classify non-fraudulent transactions correctly and with less misclassification error at the same time. Moreover, recall values of 99.91% for legitimate and 98.97% for fraud show the strength of the model in identifying actual fraudulent transactions while maintaining the ability to detect transactions of both classes. Overall, the outcome validates the HGNN model's capability of representing intricate relationships and temporal variations that lead to precise and consistent fraud detection in new data. Figure 5 depicts the graphical illustration of the results on test data.

Table 6. Ablation Study Results Comparison

| Model Variant | Accuracy (%) | Precision (%) | Recall (%) | F1-Score (%) |
|---|---|---|---|---|
| MLP | 99.61 | 96.92 | 95.73 | 96.32 |
| Homogeneous GNN | 99.74 | 97.56 | 96.48 | 96.99 |
| HGNN w/o SMOTE-Tomek | 99.79 | 97.88 | 96.94 | 97.39 |
| HGNN w/o Attention | 99.86 | 98.64 | 97.61 | 98.12 |
| **Proposed HGNN** | **99.97** | **99.15** | **98.97** | **99.48** |

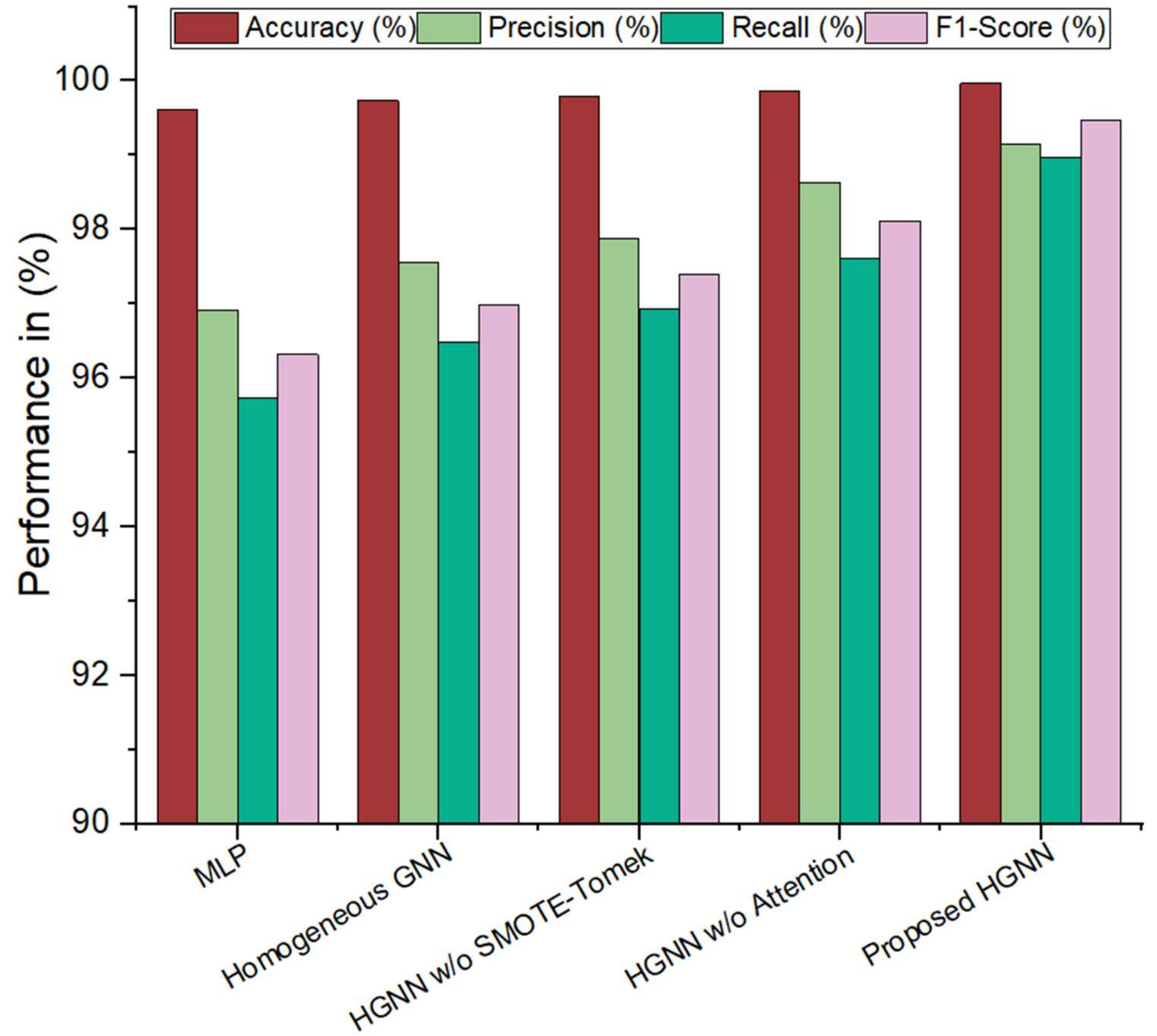

**Fig. 6 Graphical Illustration of Ablation Study Results Comparison**

The results of the ablation study comparison are depicted in Table 6, which demonstrates the impact of each model on the performance of the proposed HGNN model in a clear manner. The baseline MLP model shows the better performance on all metrics, specifically a 99.61% accuracy, 96.92% precision, 95.73% recall, and 96.32% F1-score, which underlines the drawbacks of simply classifying the transactions as separate instances without any relational modelling. The addition of the graph structure via homogeneous GNN boosted all the metrics, which indicates that the learning based on the graph improves the capturing of the transaction dependencies. The HGNN without SMOTE-Tomek shows even more improvement which implies that exclusive heterogeneous relational modelling is enough to attain better fraud discrimination and at the same time lack of imbalance handling is still limiting recall and F1-score. The HGNN without attention mechanism obtains better accuracy (99.86%), precision (98.64%), recall (97.61%), and F1-score (98.12%), thus confirming the advantage of heterogeneous graph building but at the same time highlighting the role of attention in the process of informative neighbour's selective aggregation. The proposed complete HGNN model stands out as the highest-performing one among all metrics with an impressive accuracy of 99.97%, a precision of 99.15%, a recall of 98.97%, and an F1-score of 99.48%. The use of SMOTE-Tomek for imbalance mitigation, relation-aware attention for adaptive message passing, and heterogeneous graph representation for capturing complex transactional relationships combined in the proposed HGNN model provides the best and effective fraud detection performance. Figure 6 depicts the graphical illustration of the ablation study results comparison.

Table 7. Comparison of Results with Current Models

| Models | Accuracy | Recall | Precision | F1-Score |
|---|---|---|---|---|
| Ensemble Model [11] | 93.68 | 89.20 | 98.90 | 93.80 |
| CBLOF [12] | 97.60 | 97.60 | 97.70 | 97.60 |
| AAE [13] | 99.94 | 80.08 | 82.95 | 81.49 |
| MLP [15] | 95.80 | 93.20 | 97.60 | 95.80 |
| XGBoost [17] | 99.93 | 95.00 | 91.67 | 99.30 |
| VQC-PSO [18] | 94.54 | 93.55 | 93.64 | 93.59 |
| LGBM [19] | 99.95 | 81.61 | 88.00 | 85.05 |
| DCNN [20] | 94.59 | 95.27 | 94.00 | 94.63 |
| EDCNN [20] | 95.61 | 96.62 | 94.70 | 95.65 |
| CNN-LSTM [21] | 99.93 | 89.27 | 90.51 | 89.88 |
| Ensemble ML [22] | 99.95 | 85.71 | 87.50 | 86.59 |
| Proposed Model | 99.97 | 98.97 | 99.15 | 99.48 |

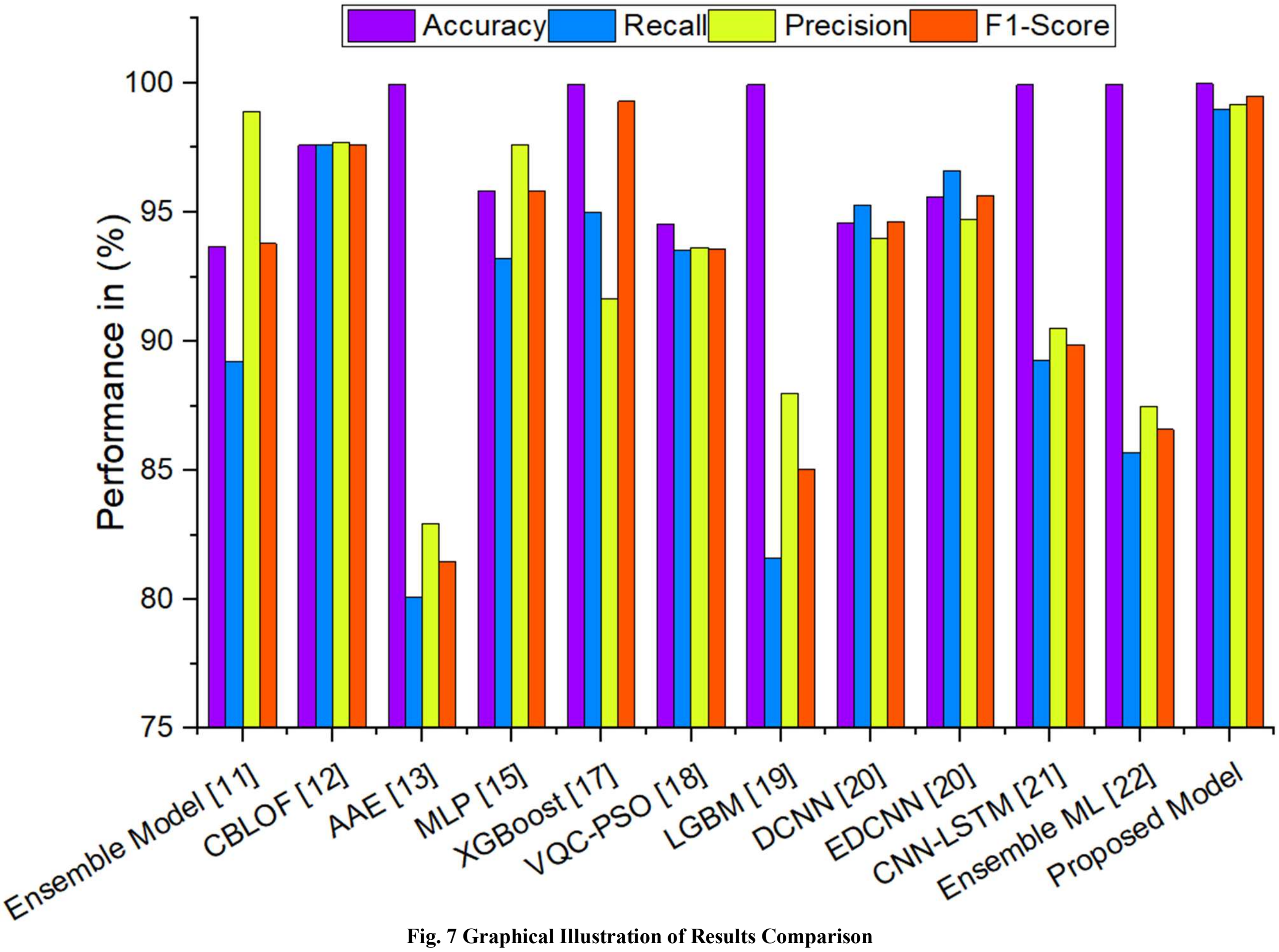

Fig. 7 Graphical Illustration of Results Comparison

In the comprehensive comparison provided in Table 7, the proposed HGNN-based fraud detection model is compared with several existing state-of-the-art methods, which clearly shows the superiority of the proposed approach. Conventional machine learning and ensemble models, including the Ensemble Model [11], MLP [15], and VQC-PSO [18], present moderate to high accuracy but get comparatively lower recall or F1-scores, thereby demonstrating the inability of these models to reliably uncover frauds to the same extent. Unsupervised and semi-supervised methods, such as CBLOF [12] and AAE [13], achieved competitive accuracy but their considerably lower recall and F1-scores show that these methods tend to analyze many fraud cases, which is unacceptable in case of financial applications. Tree-based and boosting algorithms like XGBoost [17] and LGBM [19] have shown very high accuracy, however, their relatively low recall still indicates that the models are biased towards the legal transactions and fraud detection is neglected.

DL methods such as DCNN, EDCNN, and CNN–LSTM led to improvements in recall consistency, though they still showed trade-offs between precision and recall resulting in suboptimal F1-scores. On the other hand, the proposed HGNN model has attained the most outstanding performance across the board, registering an accuracy of 99.97%, a recall of 98.97%, a precision of 99.15%, and an F1-score of 99.48%, indicating a detection ability that is not only robust but also well-balanced. Figure 7 depicts the graphical illustration of the developed HGNN model's results comparison with current methods. The results obtained from this study support the assertion that the proposed model's capability of simultaneously capturing diverse transactional relationships, class imbalance alleviation and attention-based message aggregation application leads to better generalization and substantial superiority over the existing models when evaluated using all the metrics.

### *4.4. Advantages & Limitations*

The proposed HGNN model has several advantages for CCFD. Initially, through representing transactions as a heterogeneous graph, the model successfully reflects the intricate connections between the entities like cards, transactions, time periods, and feature groups compared to other ML models. The integration of attention mechanisms makes possible the neighbor aggregation that is adaptable to the model selection of the most informative transactional interactions and thus enhances the detection accuracy. The application of SMOTE-Tomek sampling considerably reduces class imbalance resulting in high recall and F1-score for fraudulent transactions with minimal false positives. The model shows a great ability to generalize that can be seen in its regular performance on the training and testing data which makes it ready for real-time application. Overall, the proposed HGNN model surpassed the existing methods at all evaluation metrics, thus providing the confirmation of its robust and dependability for the detection of financial fraud.

Even though the proposed HGNN model shows excellent performance, it still has some limitations. The heterogeneous graph has high computational complexity and memory costs, which can restrict the scalability of the application in the case of large-scale transaction streams. The results of the HGNN model is also dependent on choices related to the design of the graph like types of nodes, definitions of edges, and temporal window selection, which need to be optimally tuned. In addition, the training process is more complicated than that of traditional models, which can lead to an increase in the effort required for implementation.

## 5. Conclusion

This study developed an efficient methodology for detecting CC fraud using HGNN, which was specifically designed to handle the difficulties of the highly unbalanced classes and the complicated transactions that are characteristic of actual financial data. The proposed method was able to depict various transactions using a heterogeneous graph architecture and by using an attention-based message passing technique, it managed to take into account the complex relationships, time factors, and user behavior that are usually neglected by the regular ML and DL approaches. The integration of SMOTE-Tomek sampling in the model further boosted its capacity to identify rare fraudulent transactions, meanwhile lowering the rate of false positives. The experiment analysis on the credit card fraud dataset showed that the HGNN model outperformed the others in comparison, reaching the highest accuracy of 99.97%, precision of 99.15%, recall of 98.97%, and F1-score of 99.48%. Through ablation study analysis, it was validated that the three individual factors: heterogeneity, attention mechanisms, and imbalance handling each contributed to the performance gains. The results demonstrated the robust, generalization ability and the real-time use of the proposed HGNN model thus confirming its potential as a solution for the intelligent fraud detection systems of the future in finance applications.

In future, the proposed framework can be extended to use real-time and streaming transaction data to enable online fraud detection with dynamic graph updates. Combining the temporal graph neural networks and self-supervised or contrastive learning strategies to lessen reliance on labelled data can further enhance the HGNN framework. The exploration of the explainable AI methods in the HGNN structure can make the model more transparent and hence more compliant with the regulations in the financial sector.

## Code Availability

The implementation of the proposed GraphFEN pipeline, including data preprocessing, heterogeneous graph construction, and evaluation code, is publicly available as a versioned Jupyter Notebook release archived on Zenodo (https://doi.org/10.5281/zenodo.18618196), with the development repository hosted on GitHub (https://github.com/kathiresan-jayabalan/graphfen-ccfd).

## Conflicts of Interest

The authors declare that there is no conflict of interest regarding the publication of this paper.

## Funding Statement

Not Applicable.